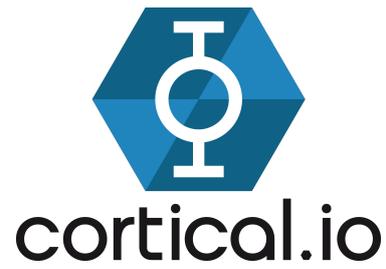

# Semantic Folding

*Theory and its Application in Semantic Fingerprinting*

White Paper

Version 1.2

Author: Francisco E. De Sousa Webber

Vienna, March 2016

# Contents

# List of Figures

# About this Document

## Evolution of this Document

| Document Version | Date | Edits | Author(s) |
|---|---|---|---|
| 1.0 | 25. November 2015 | First Public Release | Francisco Webber |
| 1.1 | 01. March 2016 | Updated sections | Francisco Webber |

## Contact

Your feedback is important to us. We invite you to contact us at the following address with your comments, suggestions or questions: info @ cortical . io (please remove spaces).

# Abstract

*The two faculties - making analogies and making predictions based on previous experiences - seem to be essential and could even be sufficient for the emergence of human-like intelligence.*

It is common scientific practice to approach phenomena, that cannot be scientifically explained by an existing set of scientific theories, through the use of statistical methods. This is how medical research led to coherent treatment procedures that are useful for patients. By observing many cases of a disease and by identifying and taking account of its various cause and effect relationships, the statistical evaluation of these records enabled the making of well thought out predictions and (consequently) the discovery of adequate treatments and countermeasures. Nevertheless, since the rise of molecular biology and genetics, we can observe how medical science is moving from the time-consuming trial and error strategy to a much more efficient, deterministic procedure that is grounded on solid theories and will eventually lead to a fully personalized medicine.

The science of language has had a very similar development. In the beginning, extensive statistical analyses led to a good analytical understanding of the nature and functioning of human language and culminated in the discipline of linguistics. Following the increasing involvement of computer science in the field of linguistics, it turned out that the observed linguistic rules were extremely hard to use for the computational interpretation of language. In order to allow computer systems to perform language based tasks comparable to humans, a computational theory of language was needed and, as no such theory was available, research again turned towards a statistical approach by creating various computational language models derived from simple word count statistics. Despite initial successes, statistical *Natural Language Processing (NLP)* still suffers from two main flaws: The achievable precision is always lower than that of humans and the algorithmic frameworks are chronically inefficient.

*Semantic Folding Theory (SFT)* is an attempt to develop an alternative computational theory for the processing of language data. While nearly all current methods of processing natural language based on its meaning use word statistics, in some form or other, Semantic Folding uses a neuroscience-rooted mechanism of distributional semantics.

After capturing a given semantic universe of a reference set of documents by means of a fully unsupervised mechanism, the resulting semantic space is folded into each and every word-representation vector. These vectors are large, sparsely filled binary vectors. Every feature bit in this vector not only corresponds to but also equals a specific semantic feature of the folded-in semantic space and is therefore semantically grounded.

The resulting word-vectors are fully conformant to the requirements for valid word-SDRs (Sparse Distributed Representation) in the context of the *Hierarchical Temporal Memory (HTM)* theory of Jeff Hawkins. While HTM theory focuses on the cortical mechanism for identifying, memorizing and predicting reoccurring sequences of SDR patterns, Semantic Folding theory describes the encoding mechanism that converts semantic input data into a valid SDR format, directly usable by HTM networks.

The main advantage of using the SDR format is that it allows any data-items to be directly compared. In fact, it turns out that by applying Boolean operators and a similarity function, many Natural Language Processing operations can be implemented in a very elegant and efficient way.

Douglas R. Hofstadter's *Analogy as the Core of Cognition* is a rich source for theoretical background on mental computation by analogy. In order to allow the brain to make sense of the world by identifying and applying analogies, all input data must be presented to the neo-cortex as a representation that is suited to the application of a distance measure.

# Part 1: Semantic Folding

## Introduction

Human language has been recognized as a very complex domain for decades. No computer system has so far been able to reach human levels of performance. The only known computational system capable of proper language processing is the human brain. While we gather more and more data about the brain, its fundamental computational processes still remain obscure. The lack of a sound computational brain theory also prevents a fundamental understanding of Natural Language Processing. As always when science lacks a theoretical foundation, statistical modeling is applied to accommodate as much sampled real-world data as possible.

A fundamental yet unsolved issue is the actual representation of language (data) within the brain, denoted as the *Representational Problem*.

Taking *Hierarchical Temporal Memory* (HTM) theory, a consistent computational theory of the human cortex, as a starting point, we have developed a corresponding theory of language data representation: The Semantic Folding Theory.

*The process of encoding words, by using a topographical semantic space as a distributional reference frame into a sparse binary representational vector is called **Semantic Folding** and is the central topic of this document.*

Semantic Folding describes a method of converting language from its symbolic representation (text) into an explicit, semantically grounded representation that can be generically processed by HTM networks. As it turns out, this change in representation, by itself, can solve many complex NLP problems by applying Boolean operators and a generic similarity function like Euclidian Distance.

Many practical problems of statistical NLP systems, like the high cost of computation, the fundamental incongruity of precision and recall[1], the complex tuning procedures etc., can be elegantly overcome by applying Semantic Folding.

---

[1] The more you get of one, the less you have of the other.

## Origins and Goals of Semantic Folding Theory

Semantic Folding Theory is built on top of Hierarchical Temporal Memory Theory[i]. The HTM approach to understanding how neo-cortical information processing works, while staying closely correlated to biological data, is somewhat different from the more mainstream projects that have either a mainly *anatomic* or a mainly *functional mapping* approach.

Neuroscientists working on micro-anatomic models [ii] have developed sophisticated techniques for following the actual 3D structure of the cortical neural mesh down to the microscopic level of dendrites, axons and their synapses. This enables the creation of a complete and exact map of all neurons and their interconnections in the brain. With this wiring diagram they hope to understand the brains functioning from the ground up.

Research in functional mapping, on the other hand, has developed very advanced imaging and computational models to determine how the different patches of cortical tissue are interconnected to form functional pathways. By having a complete inventory[iii] of all existing pathways and their functional descriptions, the scientists hope to unveil the general information architecture of the brain.

In contrast to these primary data-driven approaches, HTM-Theory aims to understand and identify principles and mechanisms by which the mammalian neo-cortex operates. Every characteristic identified can then be matched against evidence from neuro-anatomical, neuro-physiological and behavioral research. A sound theory of the neo-cortex will in the end fully explain all the empirical data that has been accumulated by generations of neuroscientists to date.

Semantic Folding Theory tries to accommodate all constraints defined by Hawkins' cortical learning principles while staying biologically plausible and explaining as many features and characteristics of human language as possible.

*SFT provides a framework for describing how semantic information is handled by the neo-cortex for natural language perception and production, down to the fundamentals of semantic grounding during initial language acquisition.*

This is achieved by proposing a novel approach to the representational problem, namely the capacity to represent meaning in a way that it becomes computable by the cortical processing infrastructure. The possibility of processing language information at the level of its meaning will enable a better understanding of the nature of intelligence, a phenomenon closely tied to human language.

## The Hierarchical Temporal Memory Model

The HTM Learning Algorithm is part of the HTM model developed by Jeff Hawkins. It is not intended to give a full description of the HTM model here, but rather to distill the most important concepts in order to understand the constraints within which the Semantic Folding mechanism operates.

### Online Learning from Streaming Data

From an evolutionary point of view, the mammalian neo-cortex is a recent structure that improves the command and control functions of the older (pre-mammalian) parts of the brain. Being exposed to a constant stream of sensorial input data, it continuously learns about the characteristics of its surrounding environment, building a sensory-motor model of the world that is capable of optimizing an individual's behavior in real time, ensuring the well-being and survival of the organism. The optimization is achieved by using previously experienced and stored information to modulate and adjust the older brain's reactive response patterns.

### Hierarchy of Regions

The neo-cortex, in general, is a two-dimensional sheet covering the majority of the brain. It is composed of microcircuits with a columnar structure, repeating over its entire extent.

Regardless of their functional role (visual, auditory or proprioceptive), the microcircuits do not change much of their inner architecture. This micro-architecture is even stable across species, suggesting that it is not only older on an evolutionary scale than the differentiation of the various mammalian families but also that it is implementing a basic algorithm to be used for all (data) processing.

Although anatomically identical, the surface of the neo-cortex is functionally subdivided into different regions. Every region receives inputs either originating from a sensorial organ or being generated by the outputs of another region. The different regions are organized in hierarchies. Every region outputs a stable representation for each learned sequence of input patterns, which means that the fluctuations of input patterns become continuously slower while ascending hierarchical layers.

## Sequence Memory

Every cortical module performs the same fundamental operations while its inputs are exposed to a continuous stream of input data among which it detects and memorizes the reoccurring sequences of patterns. Every recognized input-sequence generates a distinct output-pattern that is exposed at the output stage of the module. During the period where the input flow is within a known sequence, each module also generates a prediction pattern containing a union of all patterns that are expected to follow the current one, according to its stored sequence (experience).

The above capabilities describe a memory system rather than a processing system as one might expect to find in this highest brain structure. This memory system is capable of processing data just by storing it. What is special for this memory is that its data-input is different from its data-output and specific sequences of several input-patterns lead to a single specific output-pattern.

In contrast, all electronic memory components we integrate into computer systems, use discrete addresses to store and retrieve data over a combined input/output port. These memory-addresses do not represent data by themselves but rather denote the location of a specific storage cell within a uniformly organized array of such cells. This storage location is also independent of any actual data and holds whatever value is stored there without giving any indication on what the stored data corresponds to. All semantic information about the data has to be contributed by the associated program, that stores and reads the data values in every storage cell it uses. This indirection decouples the semantics of the data from its actual (scalar) value. It remains the sole responsibility of the associated software to handle the data in a meaningful way by executing a sequence of rules and transformations that act on the stored (scalar) data. The number of processing steps can vary substantially depending on the intended semantic goal and all intended semantic aspects have to be known in advance, at the time the program is created. The inability to predict execution latency and the need to conceptualize all semantic processing steps in advance make it hard (if not impossible) to perform the input- to output-data conversion in real-time. Furthermore, unexpected input data very often leads to a crash of the applied software.

The only technical memory architecture that comes close to the cortical memory described above is that of *Content Addressable Memory*, which corresponds in principle to standard memory cells with an address input and a data in/output, but with the exception that actual data is directly interpreted as address and fed on the address input (hence content addressable). As an example lets assume the string: '2+3'. If each of the three characters represents an 8-bit ASCII code we can interpret the string as a 24-bit address pointing to a specific memory location in an array of $2^{24}$ (approx. 16 Million) cells, in which the result '5' is stored. Whenever a term like '2+3" is entered on the address input, the result is returned one single cycle later on the data-output.

Although this seems efficient in processing terms, such an architecture needs vast amounts of memory to become a general purpose processing mechanism. A query like: "which is the highest mountain on earth" would assume an address space of 38x8-bits=304 bits assuming an array of $3.26 \times 10^{91}$ memory cells.

The fact that memory has been, for a long time, the most expensive part of a computer has dwarfed the use of CAMs to very specific and small applications like in network appliances, where the address space is small and real-time processing important.

In contrast the cortical memory has a mechanism to generate specific addresses for every memory cell depending on the data to be stored. This means that not all thinkable memory locations have to be present but that the address of a memory cell is learned whenever data gets stored. As a result, the cortical memory function implements best of both worlds: indirection-free (no processor needed) semantic data storage with a minimum number of actually implemented memory cells. The result of any "query" can always be determined in a single step while allowing query widths of thousands if not millions of bits.

This constant and minimal processing delay, independent of the nature of the processed data, is essential to provide an organism with useful real-time information about its surroundings.

A second big advantage of the HTM-CAM principle is that the amount of data that can be handled in real time increases linearly with the amount of available memory-modules. More modules mean more processing power, which is a very effective way for evolution to adapt and improve the mammalian brain: just by augmenting the amount of cortical real estate.

## Sparse Distributed Representations

The memory needed for a CAM can be substantially reduced if compression is applied to the addresses generated. Technical CAM implementations use dense binary hash values where every combination of address-bits points to a single memory location. Unfortunately, the computational effort to encode and decode the hash values is very high and counteracts - with growing memory space - the speed advantages of the CAM approach.

The architectural limitation to smaller and constant word sizes (8, 16, 32, 64 ... bits), corresponding to a dense representation scheme, became the fundament of standard computer technology and triggered the race for ever increasing clock rates pacing more and more powerful serial processing cores.

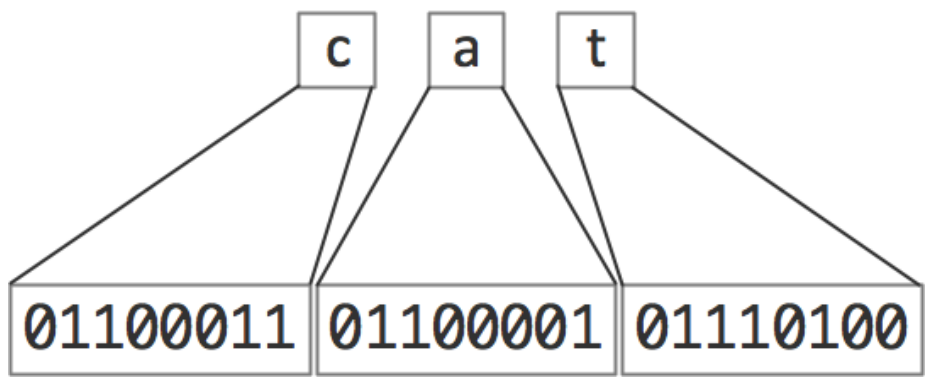

Fig. 1: Dense representation of the word *cat* - individual bits don't carry meaning

By using a dense representation format, every combination of bits identifies a specific data item. This would be efficient in the sense that it would allow for much smaller word sizes but it would also create the need for a dictionary to keep track of all the data items recorded. The longer the dictionary list would become, the longer it would take to find and retrieve any specific item. This dictionary could link the set of stimuli corresponding to the word *cat* to the identifier *011000110110000101110100* therefore materializing the semantic grounding needed to process the data generated by the surrounding world.

Instead of realizing the semantic grounding through the indirection of a dictionary, it could also occur at the representation level directly. Every bit of the representation could correspond to an actual feature of the corresponding data item that has been perceived by one or more senses. This leads to a much longer representation in terms of number of bits but these long binary words have only very few set bits (sparse filling).

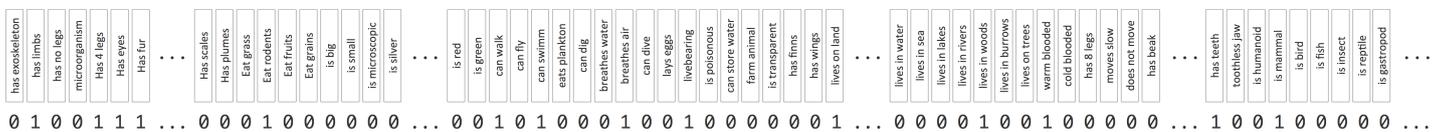

**Fig. 2: Excerpt of a sparse representation of *cat* - every bit has a specific meaning**

By storing only the positions of the set bits, a very high compression rate becomes possible. Furthermore, the use of a constantly growing dictionary for semantic grounding can be avoided.

By using a sparse data representation, CAM-computing becomes possible by simply increasing the number of cortical modules deployed. But one big problem remains: noise. Unlike silicon based devices, biological systems are very imprecise and unreliable, introducing high levels of noise into the memory-computing process. False activation or false dropping of a single bit in a dense representation renders the whole word into something wrong or unreadable. Sparse representations are more resistant to dropped bits as not all descriptive features are needed to identify a data item correctly. But shifted bit-positions are still not tolerated as the following example shows.

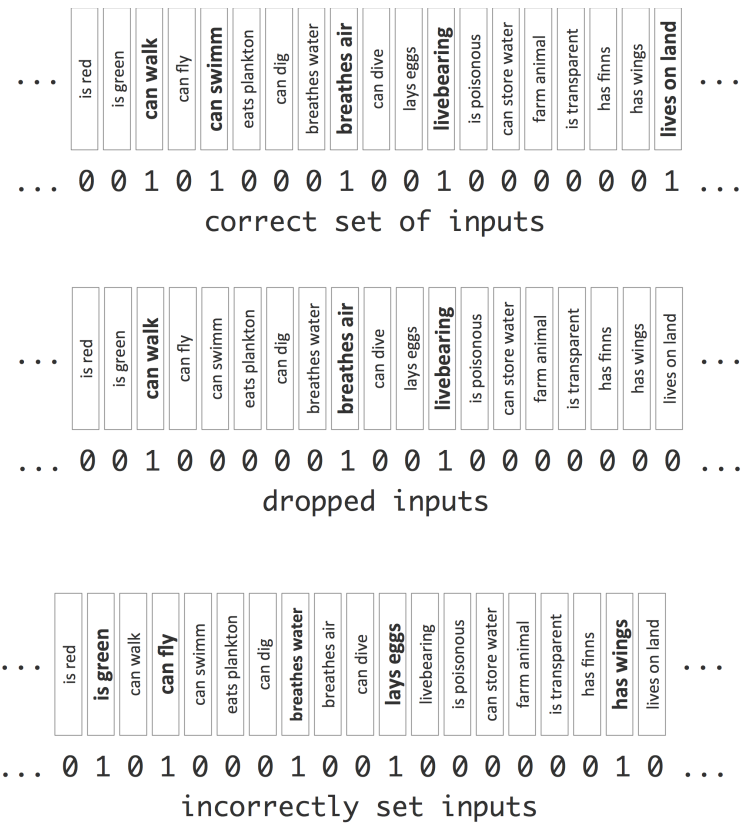

**Fig. 3: Influence of dropped and shifted bits on sparse representations**

In the above example, the various binary features are located at random positions within the sparse binary data word. A one-to-one match is necessary to compare two data-items. If we now introduce a mechanism that tries to continually group[2] the feature-bits that fire simultaneously within the data word, we gain several benefits.

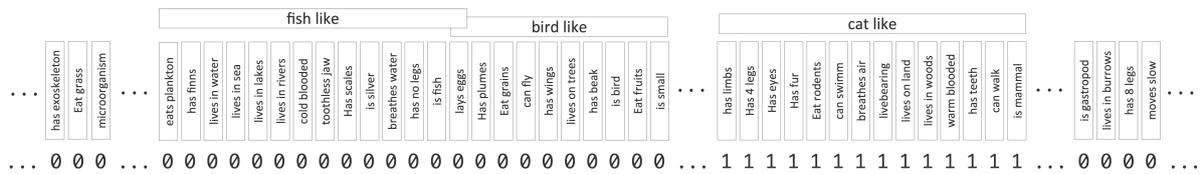

**Fig. 4: Grouping co-occurring features together improves noise resistance**

A first advantage is the substantial improvement of noise resistance in the representation of messy real-world data. When a set bit shifts slightly to the left or the right - a blur-effect happening frequently when biological building blocks are used - the semantic meaning of the whole data-word remains very stable, thus contributing only a very small error value.

---

[2] In a spatial, topographical sense.

A second advantage is the possibility to compute a gradual similarity value, allowing a much finer-grained semantic comparison, which is mandatory for functions like disambiguation and inference.

*If we assume the neo-cortex to be a memory system able to process data in real-time and to be built out of repeating microcircuits, the **Sparse Distributed Representation** is the minimum necessary data configuration, while being the biologically most convenient data format to be used.*

### Properties of SDR Encoded Data

I. **SDRs can be efficiently stored** by only storing the indices of the (very few) set bits. The information loss is negligible even if subsampled.

II. **Every bit in a SDR has semantic meaning** within the context of the encoding sensor.

III. **Similar *things* look similar,** if encoded as a SDR. Similarity can be calculated using computationally simple distance measures.

IV. **SDRs are fault tolerant** because the overall semantics of an item are maintained even if several of the set bits are discarded or shifted.

V. The **union of several SDRs** results in a SDR that still contains all the information of the constituents and behaves like a generic SDR. By comparing a new unseen SDR with a union-SDR, it can be determined if the new SDR is part of the union.

VI. SDRs can be brought to any level of sparsity in a semantically consistent fashion by using a locality based weighting scheme.

### On Language Intelligence

A frequent assumption about machine intelligence is that, in a machine executing a sufficiently complex computational algorithm, intelligence would emerge and manifest itself by generating output indistinguishable from that of humans.

In HTM theory, however, intelligence seems to be rather a principle of operation than an emerging phenomenon. The learning algorithm in the HTM microcircuits is a comparably simple storage mechanism for short sequences of SDR-encoded sensor- or input-data. Whenever a data item is presented to the circuit, a prediction of what data items are expected next is generated. This anticipatory *sensor-data* permits the pre-selection and optimization of the associated response by choosing from a library of previously experienced and stored output-SDR-sequences. This intelligent selection step is carried out by applying prediction and generalization functions to the SDR memory cells.

It has been shown that, on the one hand, prediction SDRs are generated by creating an "OR" (union) of all the stored SDRs that belong to the currently active sequence. This prediction SDR is passed down the hierarchy and used to disambiguate unclear data, to fill up incomplete data and to strengthen the storage persistence of patterns that have been predicted correctly. On the other hand, the output-SDRs can be regarded as a form of "AND" (intersection) of the stored SDRs within the current sequence. This generalized output-SDR is passed up the hierarchy to the inputs of the next higher HTM-layer leading to an abstraction of the input data to detect and learn higher-level sequences or to locate similar SDR-sequences.

In fact, intelligence is not solely rooted in the algorithm used. A perfectly working HTM circuit would not exhibit intelligent behavior by itself but only after having been exposed to sufficient amounts of relevant *special case experiences*. Neo-cortical intelligence seems to be continuously saved into the HTM-system driven by an input data stream while being exposed to the world.

## A Brain Model of Language

By taking the HTM theory as a starting point, we can characterize *Semantic Folding* as a data-encoding mechanism for inputting language semantics into HTM networks.

Language is a creation of the neo-cortex to exchange information about the semantics of the world, between individuals. The neo-cortex encodes mental concepts (stored semantics) into a symbolic representation that can be sent to muscle systems to form the externalization of the inner semantic representation. The symbols are then materialized as acoustic signals to become speech or as writing to become text or as other more exotic encodings like Morse or Braille. In order to conceive the semantics of a communication, the receiver has to convert the symbols back into the inner representation format that can than be directly utilized.

From the semantic point of view, the smallest unit[3] that contains useful, namely lexical, information consists in words.

### The Word-SDR Layer

*Hypothesis: All humans have a language receptive brain region characterized as a* **word-SDR layer***.*

During language production, language is encoded for the appropriate communication channel like speech, text or even Morse code or Braille. After the necessary decoding steps during the receiver's perception, there must be a specific location in the neo-cortex where the inner representation of a word appears for the first time.

---

[3] From a more formal lexical semantic point of view, the morpheme is the smallest unit encapsulating meaning. Nevertheless, the breaking down of words into morphemes seems to occur only after a sufficient number of word occurrences has been assimilated and probably occurs only at a later stage during language acquisition. Words would therefore be the *algorithm-generic* semantic atoms.

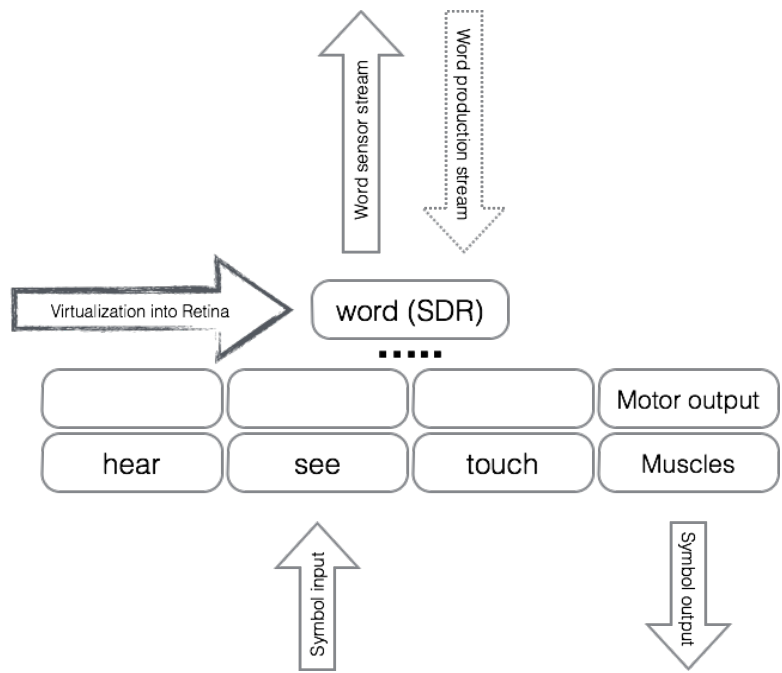

**Fig. 5: The word-SDR Hypothesis**

The brain decodes language by converting the symbolic content of phoneme-sequences or text strings into a semantically grounded neural representation of the meaning(s) of a word, the "semantic atom". These encoding and decoding capabilities are independent from the actual semantic processing. Humans are capable of learning to use new communication channels such as Braille or Morse, and can even be trained to use non-biological actuators like buttons or keyboards operated by fingers, lips, tongue or any other cortically controlled effector.

There is a place in the neo-cortex where the neurological representation of a word meaning, appears for the first time by whatever means it has been communicated. According to the HTM theory, the word representation has to be in the SDR format, as all data in the neo-cortex has this format. The word-SDRs all appear as the output of a specific hierarchical word-SDR layer. The word-SDR layer is the first step in the hierarchy of semantic processing within the neo-cortex and could be regarded as the language semantic receptive region.

Language is regarded as an inherently human capacity. No other mammal[4] is capable of achieving an information density comparable to that of human communication, which suggests that the structure of the human neo-cortex is different, in that aspect, from other mammals. Furthermore, all humans (except for rare disabilities) have the innate capability for language and all languages have very common structures. Therefore, language capacity has to be deeply and structurally rooted in the human neo-cortical layout.

---

[4] Only mammals have a neo-cortex.

## Mechanisms in Language Acquisition

Although there is much discussion about the question whether language capacity is innate or learned, the externalization of language is definitely an acquired skill, as no baby has ever spoken directly after birth.

Language acquisition is typically bootstrapped via speech and is typically extended during childhood to its written form.

## The Special Case Experience (SCE)

The neo-cortex learns exclusively by being exposed to a stream of patterns coming in from the senses. Initially, a baby is exposed to repeated basic phonetic sequences corresponding to words. The mother's repeated phonetic sequences are presented as utterances, increasing in complexity with new words being constantly introduced.

According to HTM-theory, the neo-cortex detects reoccurring patterns (word-SDRs) and stores the sequences where they appear. Every word-sequence that is perceived within a short time unit corresponds to a Special Case Experience (SCE): comparable to perceiving a visual scene. In the same way that every perceived visual scene corresponds to a *special case experience* of a set of reoccurring shapes, colors and contrasts, every utterance corresponds to an SCE, corresponding to a specific word-sequence. In the case of visual scene perception, the same objects never produce the exact same retina pattern twice. In comparison, the same concepts can be expressed with language by a very large number of concrete word combinations that never seem to repeat in their same exact manner.

*The sum of the perceived special-case-experience-utterances constitutes the only source of semantic building blocks during language acquisition.*

## Mechanisms in Semantic Grounding

The process of binding a symbol, like a written or spoken word, to a conceivable meaning represents the fundamental semantic grounding of language.

If we assume that all patterns that ascend the cortical hierarchy originate from sensorial inputs, we can hypothesize that the meaning of a word is grounded in the sensorial afferences at the very moment of the appearance of the word-SDR at the word-SDR layer. Whenever a specific word is perceived as an SCE, a snapshot of some (or all) of the sensorial afferences is made and tied to the corresponding word-SDR. Every subsequent appearance of the same word-SDR generates a new[5] sensorial snapshot (state) that is AND-ed with the currently stored one. Over time, only the bits that are in common within all states remain active. These remaining bits can therefore be said to characterize the semantic grounding of that word.

The mechanism described above is suitable for bootstrapping the semantic grounding process during the initial language acquisition phase. Over time, vocabulary acquisition is not only realized using sensory afferences but also based on known words that have been learned previously. Initially, semantic grounding through sensory states is predominant, until a basic set of concepts is successfully processed; then the definition of words using known words increases and becomes the main method for assimilating new words.

## Definition of Words by Context

The mechanism of sensory semantic grounding seems to be specific to the developing neo-cortex as the mature brain depends mostly on existing words to define new ones. The *sensorial-state semantic grounding* hypothesis could even be extended by correlating the sensorial-grounding phase with the neo-cortex before its pruning period, which explains why it is so hard for adults to specify how the generic semantic grounding could have happened during their early childhood.

The mature way of linking a concept to a specific word is by using other known words. Applying the previously introduced concept of a *Special Case Experience*, the mechanism can be described as follows: a sequence of words received by the sensory system within a sufficiently short perceptive time-interval can be regarded as an instance of a Linguistic Special Case Experience, corresponding to a statement, consisting of one or more sentences. In the case of written language, this Linguistic Special Case Experience would be a text snippet representing a context. This text snippet can be regarded as a context for every word that is contained in it. Eventually, every word will get linked to more and more new contexts, strengthening its conceptual grounding. The linking occurs by OR-ing the new Special Case Experience with the existing ones, thereby increasing the number of contexts for each word.

---

[5] The subsequent word-SDR snapshots are new in the sense that they have small differences to the previously stored one and they are the same in that they have a large overlap with the previously stored one.

## Semantic Mapping

As discussed previously the representation for a word appears in the neo-cortex at the (virtual) receptive area for words. As the neo-cortex is organized as a 2-dimensional sheet, the world-layer will be organized in a 2-dimensional fashion too. While the brain is continuously exposed to incoming data and the extent of the cortical area is relatively stable, there has to be a mechanism that reuses the available storage space by storing the incoming SCEs in an associative way one on top of each other. This can be achieved by the simple mechanism of having incoming bits that are triggered simultaneously, for features that occur concurrently and which are therefore semantically related, attract each other spatially within the 2-dimensional word-layer. As a result, every word is represented by its contexts that are distributed across the word-layer and every context-position within the 2D-area of the word-layer consists of semantically highly related SCEs (utterances) that themselves share a significant number of words. After having perceived a large enough number of SCEs the context positions end up being clustered in a way that puts similar (semantically close) contexts near to each other and dissimilar ones more distant from each other.

This mechanism ensures that two brains that are continuously exposed to similar SCEs end up having their contexts grouped in a similar way, creating an implicit consensus on the semantic distribution of features without requiring to be exposed to the exact same data. This constitutes a primary language learning mechanism that captures and defines the acquired semantic space of an individual while assuring a high degree of compatibility with the semantic space of his/her peers. Every word is semantically grounded by expressing it as a binary distribution of these context features, which are themselves distributed across the learned semantic space.

The primary acquisition of the 2D-semantic space as a distributional reference for the encoding of word meaning is called Semantic Folding. In more formal terms: every word is characterized by the list of contexts in which it appears. A context being itself the list of terms encountered in a previously stored SCE (utterance or text snippet).

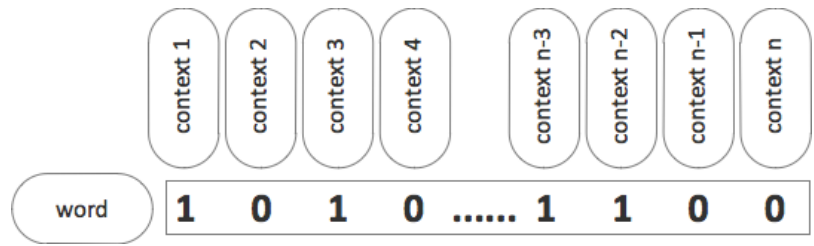

Fig. 6: Creation of a simple 1D-word-vector

This one-dimensional vector could be directly used to represent every word. But to unlock the advantages of Semantic Folding, a second mapping step is introduced that not only captures the co-occurrence information but also the semantic relations among contexts to enable *understanding* through the similarity of distributions.

Technically speaking, the contexts represent vectors that can be used to create a two-dimensional map in such a way that similar context-vectors are placed closer to each, using topological (local) inhibition mechanisms and/or by using competitive Hebbian learning principles.

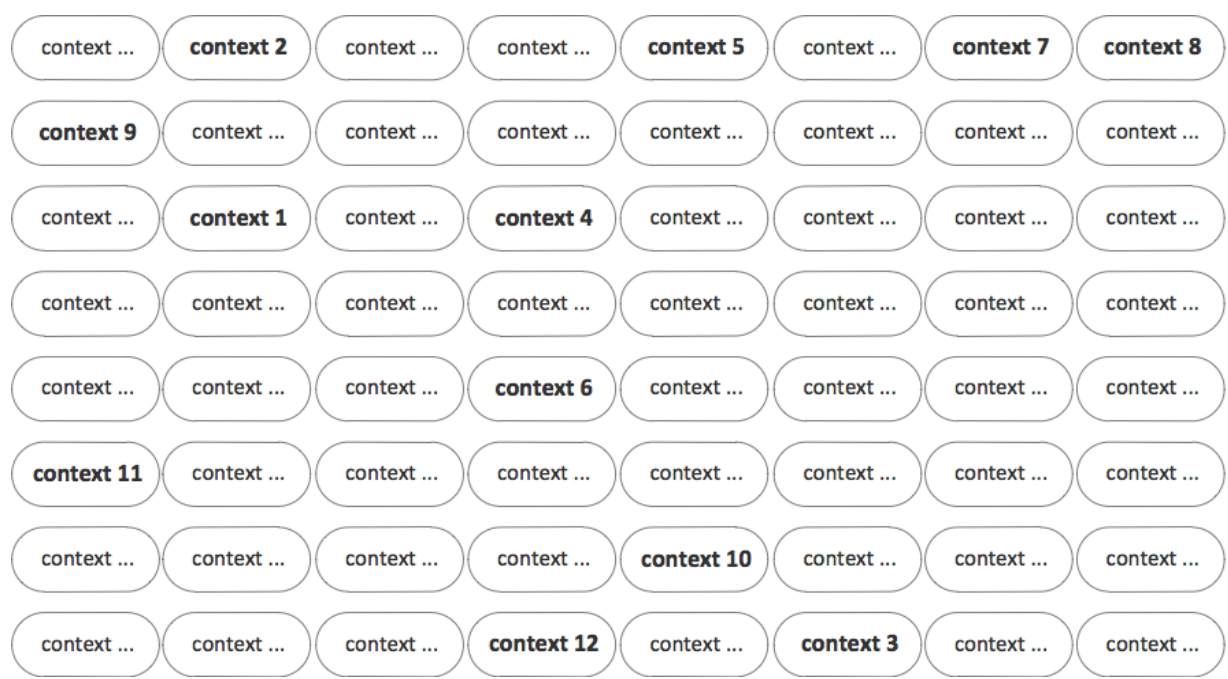

This results in a 2D-map that associates a coordinate pair to every context in the repository of contexts (the sum of all perceived SCEs). This mapping process can be maintained dynamically by always positioning a newly perceived SCE onto the map, and is even capable of growing the map on its borders if new words or new concepts appear.

*Every perceived SCE strengthens, adjusts or extends the existing semantic map.*

This map is then used to encode every single word by associating a binary vector with each word, containing a "1", if the word is contained in the context at a specific position and a "0" if not, for all positions in the map.

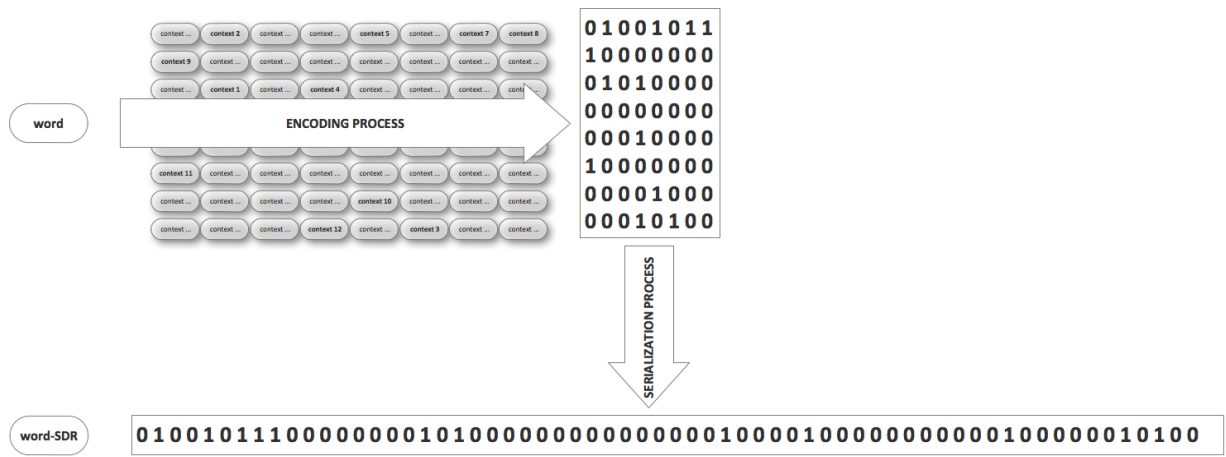

Fig. 8: Encoding of a word as word-SDR

After serialization, we have a binary vector that has a *natural* SDR format:

- A word typically appears only in a very small number of the stored contexts. The vector is therefore sparse.

- Although the vector is used in its serialized notation, the neighboring relationships between the different positions are still governed by the 2D topology, corresponding to a topological 2D-distribution.

- If a set bit shifts its position (up, down, left or right), it will misleadingly represent a different adjacent context. But as adjacent contexts have a very similar meaning due to the folded-in map, the error will be negligible or even unnoticeable representing a high noise resistance.

- Words with similar meanings look similar due to the topological arrangement of the individual bit-positions.

- The serialized word-SDRs can be efficiently compressed by only storing the indices of the set bits.

- The serialized word-SDRs can be subsampled to a high degree without losing significant semantic information.

- Several serialized word-SDRs can be aggregated using a bitwise OR function without losing any information brought in by any of the union's members.

## Metric Word Space

The set of all possible word-SDRs corresponds to a word-vector-space. By applying a distance metric (like Euclidian distance) that represents the semantic closeness of two words, the word-SDR space satisfies the requirements of a metric space:

- Distances between words are always non-negative.
- If the distance between two words is 0 then the two words are semantically identical (perfect synonyms).
- If two words *A* and *B* are in a distance *d* from each other, $d(A,B) = d(B,A)$. (Symmetry).
- For three distinct words *A, B, C* we have $d(A,C) <= d(A,B) + d(B,C)$. (Triangle inequality).

By considering the word-SDR space as a metric space, we can revert to a rich research corpus of mathematical properties, characteristics and tools that find their correspondence in the metric space representation of natural language.

## Similarity

Similarity is the most fundamental operation performed in the metric word-SDR-space. Similarity should not be directly interpreted as word synonymy, as this only represents a special case of semantic closeness that assumes a specific type of distance measure, feature selection and arrangement. Similarity should be seen as a more flexible concept that can be tuned to many different, language relevant nuances like:

- Associativity
- Generalization
- Dependency
- Synonymy
- Etc.

The actual distance measure used to calculate similarity can be varied depending on the goal of the operation. As the word-SDR vectors are composed of binary elements, the simplest distance measure consists in calculating the binary overlap. Two vectors are close if the number of overlapping bits is large. But care must be taken, as very unequal word-frequencies can lead to misinterpretations. By comparing a very frequent word that has many set-bits with a rare word having a small number of set-bits, even a full overlap would only result in the small number of overlap-bits corresponding to the number of ones in the low-frequency term.

Other distance/similarity measures that could be applied:

- Euclidian distance

- Hamming distance

- Jaccard similarity

- Cosine similarity

- Levenshtein distance

- Sørensen–Dice index

- Etc.

## Dimensionality in Semantic Folding

The Semantic Folding process takes symbolic word representation as input and converts it into an n-dimensional SDR-vector that is semantically grounded through the 2D materialization-step of the semantic map.

The main reason to choose a 2D-map over any other possible dimensionality primarily lies in the fact that the word-SDRs are intended to be fed into cortical processing systems that in turn try to implement cortical processing schemes, which happen to have evolved into a 2D arrangement.

In order to achieve actual materialization of the word-SDRs, the neighboring relationships of adjacent bits in the data should directly translate to the topological space of neo-cortical circuits. Without successful materialization of the word-SDRs, semantic grounding would not be possible, making inter-individual communication - the primary purpose of language - extremely unreliable, if not impossible.

To propagate the map-topology throughout the whole cortical extent, all afferences and efferences to or from a specific cortical area have to maintain their topological arrangement. These projections can be links between regions or pathways between sensory organs and the cortical receptive fields.

If we consider the hypothetical word-SDR layer in the human cortex to be a receptive field for language, the similarity to all other sensorial input systems, using data in a 2D arrangement, becomes obvious:

- The organ of Corti in the cochlea is a sheet of sensorial cells, where every cell transmits a specific piece of sound information depending on where on the sheet it is positioned.

- Touch is obviously generating topological information of the 2D surface of the body skin.
- The Retina is a 2D structure where two neighboring pixels have a much higher probability of belonging to the same object than two distant ones.

Topographic projection seems to be a main neuro-anatomic principle that maintains an ordered mapping from a sensory surface to its associated cortical receptive structures. This constitutes another strong argument for using a 2D semantic map for Semantic Folding.

## Language for Cross-Brain Communication

It seems reasonable to assume that a major reason for the development of language is the possibility for efficient communication. In the context currently discussed, communication can be described as the capability to send a representation of the current (neo-cortical) brain state to another individual who can then experience or at least infer the cortical status of the sender. In this case, the sensorial afferences of one neo-cortex could become part of the input of a second neo-cortex that might process this compound data differently than the sender, as it accesses a different local SCE repository. The receiving individual can then communicate this new, alternative output state back to the first individual, therefore making cooperation much easier. In evolutionary biology terms, this mechanism can be regarded as a way of extending the cortical area beyond the limits of a single subject by extending information processing from one brain to the cortical real estate of social peers. The evolutionary advantages resulting from this extended cortical processing are the same that drove the growing of the neo-cortex in the first place: higher computational power or increased intelligence by extending the overall cortical real estate available to interpret the current situational sensory input.

Although every individual uses his proprietary version of a semantic map formed along his ontogenetic development, it is interesting to note that this mechanism nevertheless works efficiently. As humans who live in the same vicinity share many genetic, developmental and social parameters, the chance of their semantic maps evolving in a very similar fashion is high: they speak the same language, therefore their semantic maps have a large overlap. The farther apart two individuals are, regardless of whether this is measured by geographical, socio-cultural or environmental distance, the smaller the overlap of their maps will be, making communication harder.

By developing techniques to record and playback language, such as writing systems, it became possible to not only make brain states available over space but also over time. This created a fast way to expose the cortex of an individual to a large set of historically accumulated Special Case Experiences (brain-states), which equally led to incremental improvement of the acquired cortical ability to make useful interpretations and predictions.

# Part 2: Semantic Fingerprinting

## Theoretical Background

HTM Theory and Semantic Folding Theory are both based on the same conceptual foundations. They aim to apply the newest findings in theoretical neuroscience to the emerging field of Machine Intelligence. The two technologies work together in a complementary fashion, Cortical.io's Semantic Folding is the encoder for the incoming stream of data, and Numenta's NuPIC (Numenta Platform for Intelligent Computing) is the intelligent backend.

Cortical.io has implemented Semantic Folding as a product called Retina API, that enables the conversion of text data into a cortex compatible representation - technically called Sparse Distributed Representation (SDR) **-** and the operation of similarity and Boolean computations on these SDRs.

### Hierarchical Temporal Memory

The Hierarchical Temporal Memory (HTM) theory is a functional interpretation of practical findings in neuroscience research. HTM theory sees the human neo-cortex as a 2D sheet of modular, homologous microcircuits that are organized as hierarchically interconnected layers. Every layer is capable of detecting frequently occurring input patterns and learning time-based sequences thereof.

The data is fed into an HTM layer in the form of Sparse Distributed Representations.

SDRs are large binary vectors that are very sparsely filled, with every bit representing distinct semantic information. According to the HTM theory, the human neo-cortex is not a processor but a memory system for SDR pattern sequences.

When an HTM layer is exposed to a stream of input data, it starts to generate predictions of what it thinks would be the next incoming SDR pattern based on what patterns it has seen so far. In the beginning, the predictions will, of course, differ from the actual data but a few cycles later the HTM layer will quickly converge and make more correct predictions. This prediction capability can explain many behavioral manifestations of intelligence.

## Semantic Folding

In order to apply HTM to a practical problem, it is necessary to convert the given input data into the SDR format. What characterizes SDRs?

- SDRs are large binary vectors (from several thousands to many millions of bits).
- SDRs have a very small fraction of their bits set to "1" at a specific point in time.
- Similar data **looks** similar when converted into SDR format.
- Every bit in the SDR has specific (accountable) meaning.
- The union of several SDRs results in an SDR that still contains all the information of its constituent SDRs.

The process of Semantic Folding encompasses the following steps:

- Definition of a reference text corpus of documents that represents the *Semantic Universe* the system is supposed to work in. The system will know all vocabulary and its practical use as it occurs in this Language Definition Corpus (LDC).
- Every document from the LDC is cut into text snippets with each snippet representing a single context.
- The reference collection snippets are distributed over a 2D matrix (e.g. 128x128 bits) in a way that snippets with similar topics (that share many common words) are placed closer to each other on the map, and snippets with different topics (few common words) are placed more distantly to each other on the map. This produces a 2D semantic map.
- In the next step, a list of every word contained in the reference corpus is created.
- By going down this list word by word, all the contexts a word occurs in are set to "1" in the corresponding bit-position of a 2D mapped vector. This produces a large, binary, very sparsely filled vector for each word. This vector is called the Semantic Fingerprint of the word. The structure of the 2D map (the Semantic Universe) is therefore *folded into* each representation of a word (Semantic Fingerprint). The list of words with their fingerprints is stored in a database that is indexed to allow for fast matching. The system that converts a given word into a fingerprint is called the Retina, as it acts as a *sensorial organ for text*. The fingerprint database is called the Retina Database (Retina DB).

## Retina DB

The Retina DB consists of actual utterances that are distributed over a 128x128 grid. At each point of the matrix we find one to several text snippets. Their constituent words represent the topic located at this position in the semantic space. The choice of implementing the semantic space as a 2D structure is in analogy to the fact that the neo-cortex itself, like all biological sensors (e.g. the retina in the eye, the Corti organ in the ear, the touch sensors in the skin etc.), is arranged as a 2 dimensional grid.

### The Language Definition Corpus

By selecting Wikipedia documents to represent the language definition corpus, the resulting Retina DB will cover general English. If, on the contrary, a collection of documents from the PubMed archive is chosen, the resulting Retina DB will cover medical English. A LDC collection of Twitter messages will lead to a "Twitterish" Retina. The same is, of course, true for other languages: The Spanish or French Wikipedia would lead to general Spanish or general French Retinas.

The size of the generated text snippets determines the *associativity bias* of the resulting Retina. If the snippets are kept very small, (1-3 sentences) the word *Socrates* is linked to *synonymous* concepts like *Plato*, *Archimedes* or *Diogenes*. The bigger the text snippets are, the more the word *Socrates* is linked to associated concepts like *philosophy*, *truth* or *discourse*. In practice, the bias is set to a level that best matches the problem domain.

### Definition of a General Semantic Space

In order to achieve cross language functionality, a Retina for each of the desired languages has to be generated while keeping the topology of the underlying semantic space the same. As a result, the fingerprint for a specific concept like *philosophy* is nearly the same in all languages (having a similar topology).

### Tuning the Semantic Space

By creating a specific Retina for a given domain, all word-fingerprints make better use of the available real estate of the 128x128 area, therefore improving the semantic resolution.

Tuning a Retina means selecting relevant representative training material. This content selection task can be best carried out by a domain expert, in contrast to the optimization of abstract algorithm parameters that traditionally require the expertise of computer scientists.

## REST API

The Retina engine, as well as an exemplary English Wikipedia Database, is available as a freely callable REST API for experimentation and testing.

A web accessible sandbox can be used by pointing a browser to http://api.cortical.io. All functionalities described in this document can be interactively tested there.

A first call to:

API: **/retina** endpoint. Get information on the available Retina will return specifics for the published Retinas.

```
[
  {
    "retinaName": "en_associative",
    "description": "An English language retina balancing synonymous
and associative similarity.",
    "numberOfTermsInRetina": 854523,
    "numberOfRows": 128,
    "numberOfColumns": 128
  }
]
```

Fig. 9: Calling the Retina API to get information on the Retina Database

# Word-SDR – Sparse Distributed Word Representation

With the Retina API it is possible to convert any given word (stored in the Retina DB) into a word-SDR. These word-SDRs constitute the *Semantic Atoms* of the system. The word-SDR is a vector of 16,384 bits (128x128) where every bit stands for a concrete context (topic) that can be realized as a *bag of words* of the training snippets at this position.

Due to the topological arrangement of the word-SDRs, similar words like *dog* and *cat* do actually have similar word-SDRs. The similarity is measured in the degree of overlap between the two representations. In contrast, the words *dog* and *truck* have far fewer overlapping bits.

## Term to Fingerprint Conversion

At the

API: **/terms** endpoint. Convert any word into its fingerprint the word *apple* can be converted into the Semantic Fingerprint that is rendered as a list of indexes of all set bits:

```
["fingerprint": {
    "positions": [

1,2,3,5,6,7,34,35,70,77,102,122,125,128,129,130,251,252,255,258,31
9,379,380,381,382,383,385,389,392,423,507,508,509,510,511,513,517,
535,551,592,635,636,637,638,639,641,643,758,764,765,766,767,768,77
1,894,900,1016,1017,1140,1141,1143,1145,1269,1270,1271,1273,1274,1
275,1292,1302,1361,1397,1398,1399,1400,1401,1402,1403,1407,1430,15
26,1527,1529,1531,1535,1655,1656,1657,1658,1659,1716,1717,1768,178
5,1786,1788,1790,1797,1822,1823,1831,1845,1913,1915,1916,1917,1918
,1931,2020,2035,2043,2044,2046,2059,2170,2176,2298,2300,2302,2303,
2309,2425,2512,2516,2517,2553,2554,2630,2651,2682,2685,2719,2766,2
767,2768,2773,2901,3033,3052,3093,3104,3158,3175,3176,3206,3286,32
91,3303,3310,3344,3556,3684,3685,3693,3772,3812,3940,3976,3978,397
9,4058,4067,4068,4070,4104,4105,4194,4196,4197,4198,4206,4323,4324
,4361,4362,4363,4377,4396,4447,4452,4454,4457,4489,4572,4617,4620,
4846,4860,4925,4970,4972,5023,5092,5106,5113,5114,5134,5174,5216,5
223,5242,5265,5370,5434,5472,5482,5495,5496,5497,5498,5607,5623,57
51,5810,6010,6063,6176,6221,6336,6783,7174,7187,7302,7427,7430,754
5,7606,7812,7917,7935,8072,8487,8721,8825,8827,8891,8894,8895,8896
,8898,8899,8951,9014,9026,9033,9105,9152,9159,9461,9615,9662,9770,
9779,9891,9912,10018,10090,10196,10283,10285,10416,10544,10545,105
86,10587,10605,10648,10649,10673,10716,10805,10809,10844,10935,109
36,11050,11176,11481,11701,12176,12795,12811,12856,12927,12930,129
31,13058,13185,13313,13314,13442,13669,14189,14412,14444,14445,144
46,14783,14911,15049,15491,15684,15696,15721,15728,15751,15752,158
33,15872,15875,15933,15943,15995,15996,15997,15998,15999,16000,161
22,16123,16127,16128,16129,16198,16250,16251,16255,16378
        ]
    }
]
```

Fig. 10: A Semantic Fingerprint in JSON format as the Retina-API returns it

The /terms endpoint also accepts multi word terms like *New York* or *United Nations* and is able to represent domain specific phrases like *Director of Sales and Business Development* or *please fasten your seat belts* by a unique Semantic

Fingerprint. This is achieved by using the desired kind of tokenization during the Retina creation process.

## Getting Context

In order to get back words for a given Semantic Fingerprint ("what does this fingerprint mean?") the following endpoint can be used:

API: **/terms/similar_terms** endpoint. Find the closest matching Fingerprint

This endpoint can be called to obtain the terms having the most overlap with the input Fingerprint. The terms with most overlap constitute the contextual terms for a specific Semantic Fingerprint.

As terms are usually ambiguous and have different meanings in different contexts, the similar terms function returns contextual terms for all existing contexts.

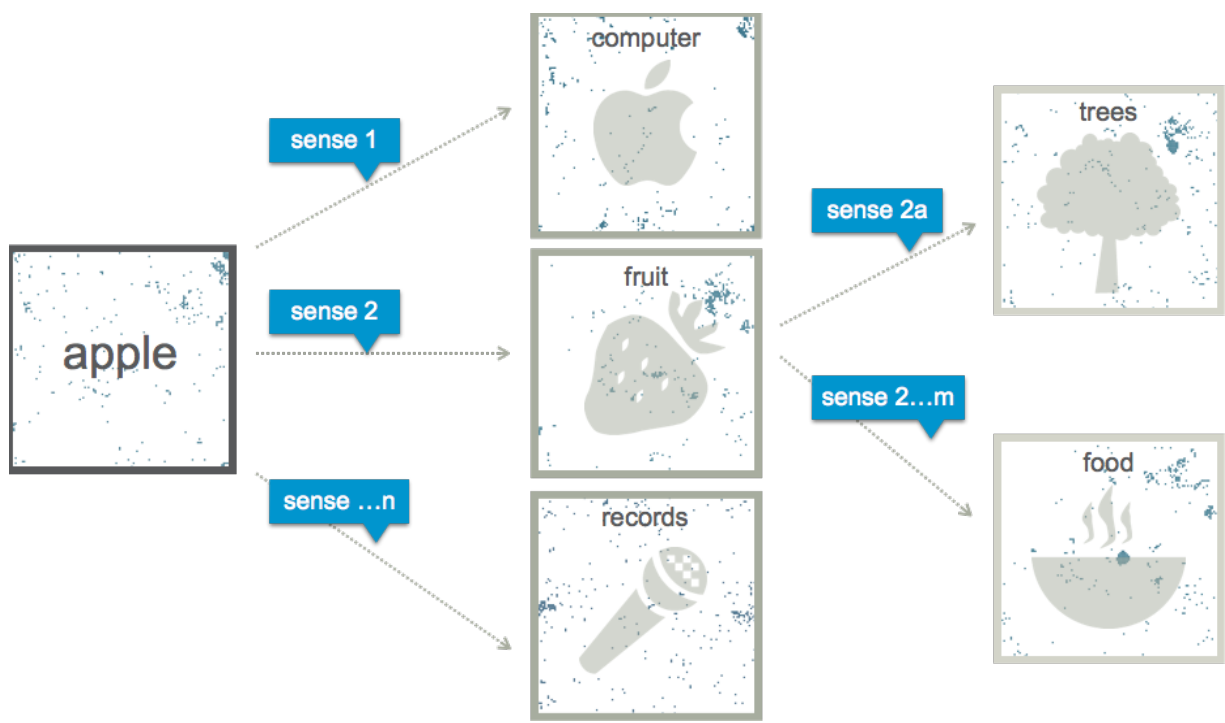

**Fig. 11 Word Sense Disambiguation of the word *apple***

The fingerprint representation of the word *apple* contains all the different meanings like computer-related meaning, fruit-related meaning or records-related meaning. If we assume the following sequence of operations:

1) get the word with the most overlap with the word *apple*: the word *computer*
2) set all bits that are shared between *apple* and *computer* to "0"

3) send the resulting fingerprint again to the similar terms function and get: the word *fruit*

4) set all bits that are shared between the fingerprint from step 2 and *fruit* to "0"

5) send the resulting fingerprint again to the similar terms function and get: the word *records*

6) ... continue until no more bits are left,

then we have identified all the contexts that this Retina knows for the word *apple*. This process can be applied to all words contained in the Retina. In fact, this form of computational disambiguation can be applied to any Semantic Fingerprint.

API: **/terms/context** endpoint. Find the different contexts of a term

Using this endpoint, any term can be queried for its contexts. The most similar contextual term becomes the label for this context and the subsequent most similar terms are returned. After having identified the different contexts, the similar term list for each of them can be queried.

## Text-SDR – Sparse Distributed Text Representation

The word-SDRs represent atomic units and can be aggregated to create document-SDRs (Document Fingerprints). Every constituent word is converted into its Semantic Fingerprint. All these fingerprints are then stacked and the most often represented features produce the highest bit stack.

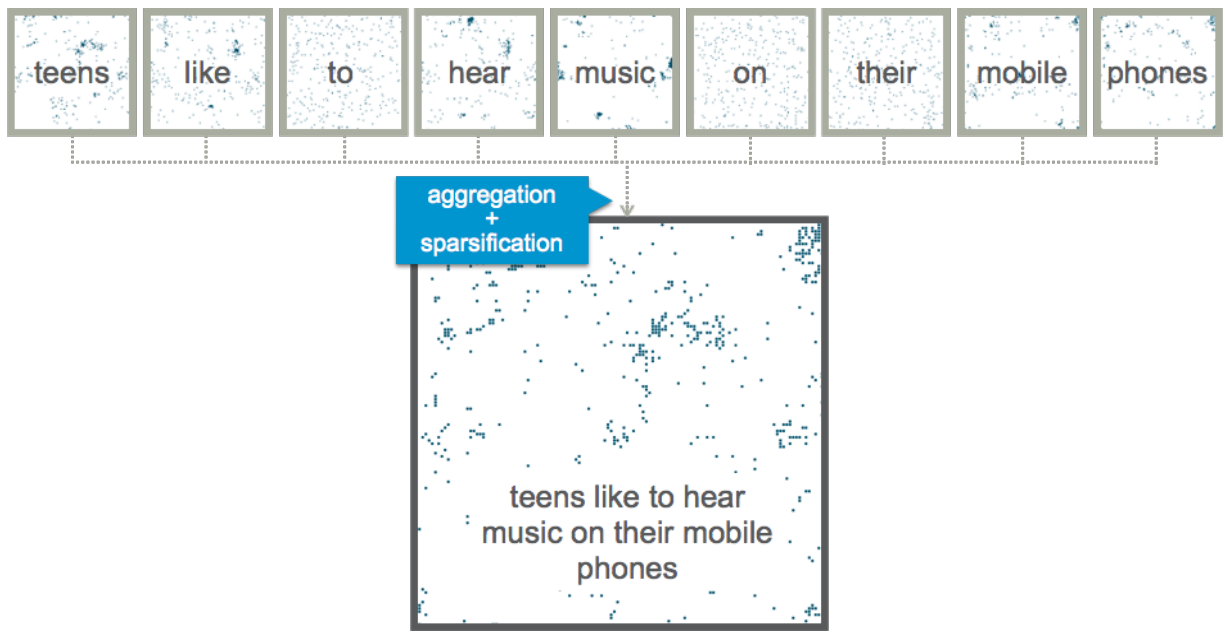

Fig. 12 Aggregation of word-SDRs into a text-SDR

## Text to Fingerprint Conversion

The bit stacks of the aggregated fingerprint are now cut at a threshold that keeps the sparsity of the resulting document fingerprint at a defined level.

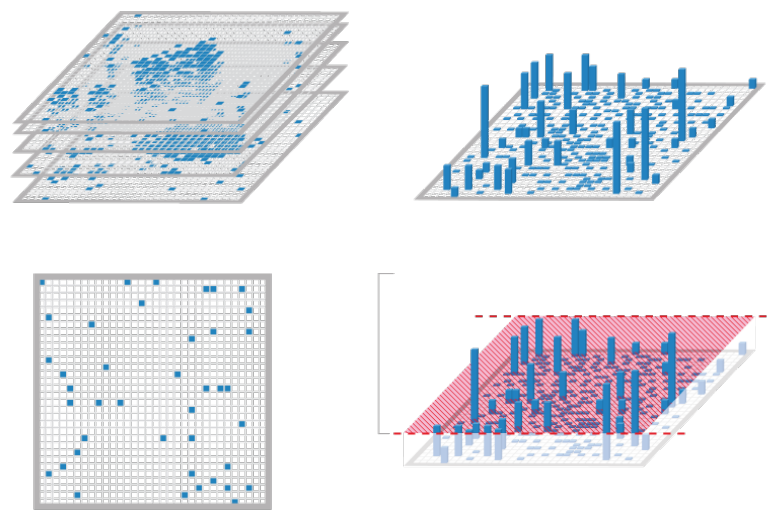

Fig. 13 The sparsification has an implicit disambiguation effect

The representational uniformity of word-SDRs and document-SDRs makes semantic computation easy and intuitive for documents of all sizes.

Another very useful side effect is that of implicit disambiguation. As previously seen, every word has feature bits for many different context groups in its fingerprint rendering. However, only the bits of the topics with the highest stacks will remain in the final document fingerprint. All other (ambiguous bits) will be eliminated during aggregation.

Fingerprints of texts can be generated using the following endpoint.

API: **/text** endpoint. Convert any text into its fingerprint

Based on the text-SDR mechanism, it is possible to dynamically allow the Retina to learn new, previously unseen words as they appear in texts.

### Keyword Extraction

The word-SDRs also allow a very efficient mechanism to extract the semantically most important terms (or phrases) of a text. After internally generating a document-fingerprint, each fingerprint of the constituent words is compared to it. The smallest set of word-fingerprints that is needed to reconstruct the document-fingerprint represents the semantically most relevant terms of the document.

API: **/text/keywords** endpoint. Find the key terms in a piece of text

### Semantic Slicing

It is often necessary to slice text into topical snippets. Each snippet should have only one (main) topic. This is achieved by stepping through the text word-by-word and sensing how many feature bits change from one sentence to the next. If many bits change from one sentence-fingerprint to the next, it can be assumed that a new topic appeared and the text is cut at this position.

API: **/text/slices** endpoint. Cutting text into topic-snippets

# Expressions – Computing with fingerprints

As all Semantic Fingerprints are homologous (they have the same size and their feature space is equal), they can be used directly in Boolean expressions. Setting and resetting selections of bits can be achieved by AND-ing, OR-ing and SUBtracting Semantic Fingerprints with each other.

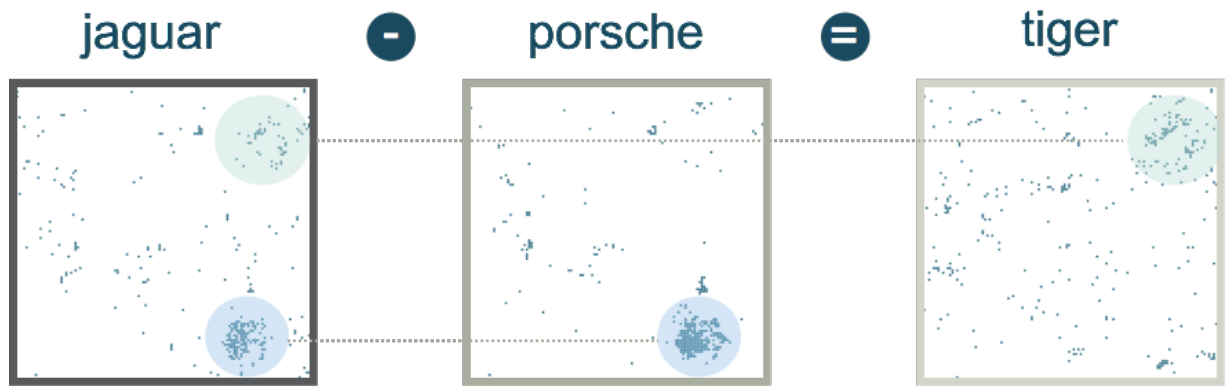

Fig. 14 Computing with word meanings

Subtracting the fingerprint of *Porsche* from the fingerprint of *jaguar* means that all the *sports car* dots are eliminated in the *jaguar* fingerprint, and that only the *big cat* dots are left. Similar but not equal would be to make an AND of the *jaguar* and the *tiger* fingerprints.

AND-ing *organ* and *liver* eliminates all *piano* and *church* dots (bits) initially also present in the *organ* fingerprint.

The expression endpoint at:

API: **/expressions** endpoint. Combining fingerprints using Boolean operators.

can be used to create expressions of any complexity.

## Applying Similarity as the Fundamental Operator

Using the Semantic Fingerprint representation for a piece of text corresponds to having semantic features in a metric space. Vectors within this metric space are compared using distance measures. The Retina API offers several different measures, some of which are absolute, which means that they only take a full overlap into account, others also take the topological vicinity of the features into account.

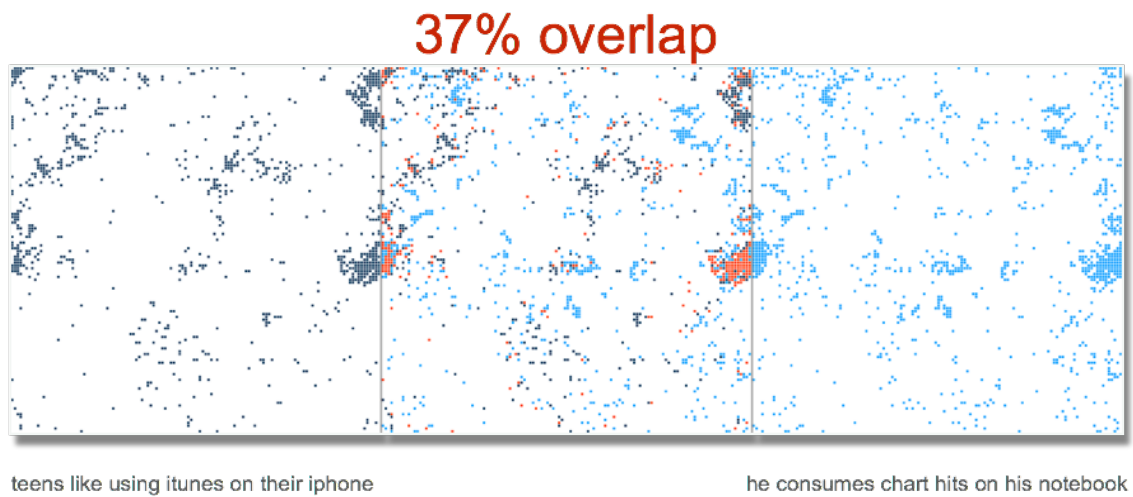

Fig. 15 Similar text snippets result in similar fingerprints

There are two different semantic aspects that can be detected while comparing two Semantic Fingerprints:

- The absolute number of bits that overlap between two fingerprints describes the semantic closeness of the expressed concepts.
- By looking at the topological position where the overlap happens, the shared contexts can be explicitly determined.

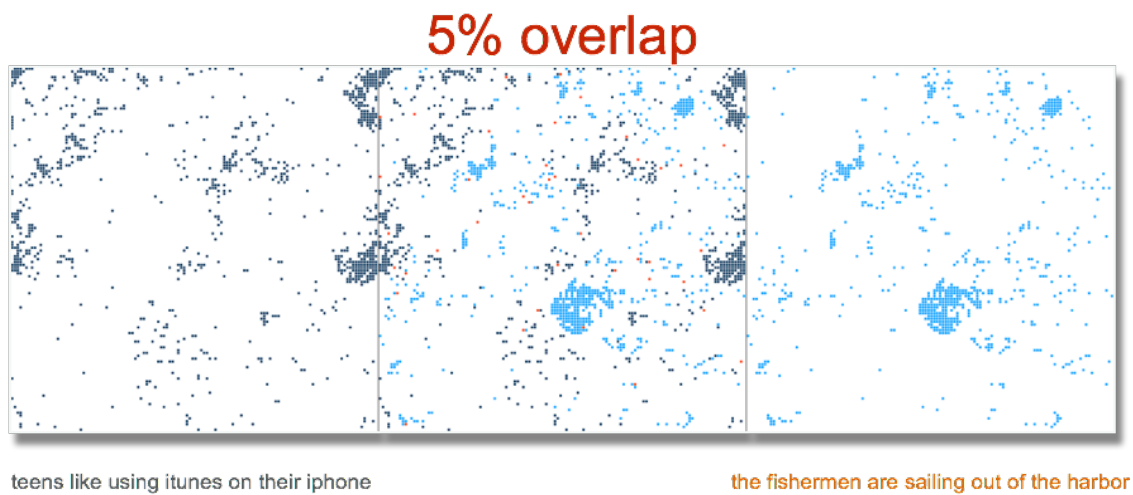

Fig. 16 Distinct text snippets result in dissimilar fingerprints

Because they are expressed through the combination of 16K features, the semantic differences can be very subtle.

## Comparing Fingerprints

API: **/compare** endpoint. Calculating the distance of two Fingerprints

The comparison of two Semantic Fingerprints is a purely mathematical (Boolean) operation that is independent of the Retina used to generate the fingerprints.
This makes the operation very fast, as only bits are compared, but also very scalable, as every comparison constitutes an independent computation and can therefore be spread across as many threads as needed to stay in a certain timing window.

## Graphical Rendering

API: **/image/compare** endpoint. Display the comparison of two Fingerprints

For convenience, image representation of Semantic Fingerprints can be obtained from the image endpoint, to be included in GUIs or rendered reports.

# Application Prototypes

Based on the fundamental similarity operation, many higher-level NLP functionalities can be built. The higher-level functions in turn represent building blocks that can be included in many different business cases.

## Classification of Documents

Traditionally, document classifiers are defined by providing a sufficiently large number of pre-classified documents and then by training the classifier with these training documents. The difficulty of this approach is that many complex classification tasks across a larger number of classes require large amounts of correctly labeled examples. The resulting classifier quality degrades in general with the number of classes and their (semantic) closeness.

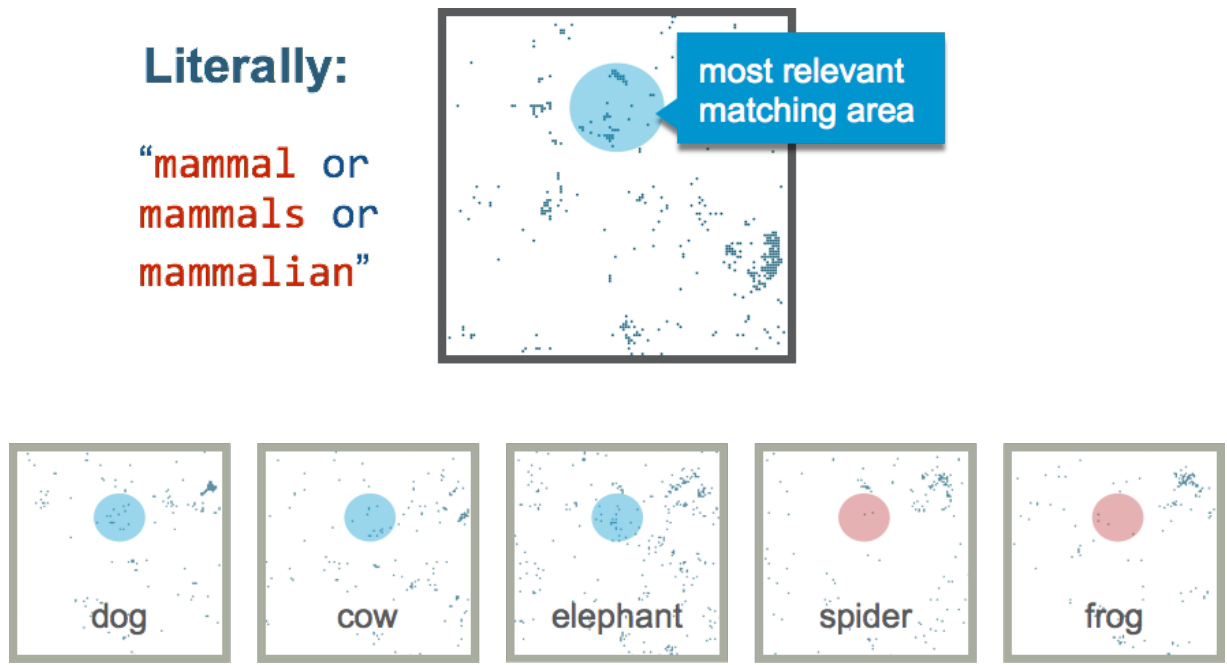

Fig. 17 Classification using Semantic Fingerprints

With Semantic Fingerprints, there is no need to train the classifier. The only thing needed is a reference fingerprint that specifies an explicit set of semantic features describing a class. This reference set or semantic class skeleton can be obtained either through direct description by enumerating a small number of generic class features and creating a Semantic Fingerprint of this list (for example the three words "mammal" + "mammals" + "mammalian"), or by formulating an expression. By computing the expression: "tiger" AND "lion" AND "panther", a Semantic Fingerprint is created that specifies *big cat* features.

For the creation of subtler classes, the classify endpoint:

API: **/classify/create_category_filter** endpoint. Create a category filter for a classifier can be used to create optimized category filters based on a couple of example documents. The creation of a filter fingerprint has close to no latency (compared to the usual classifier training process), which makes *on the fly* classification possible. The actual classification process is done by generating a text fingerprint for each document and comparing it with each category filter fingerprint. By setting the (similarity) cut-off threshold accordingly, the classification sensitivity can be set optimally for each business case. As the cutoff is specified relative to the actual semantic closeness, it does not cause any deterioration in recall.

## Content Filtering Text Streams

Filtering text streams is also done using the fingerprint classifier described before. The main difference is that the documents do not preexist but are classified as they come in. The streaming text sources can be of any kind like tweets, news, chat, Facebook posts etc.

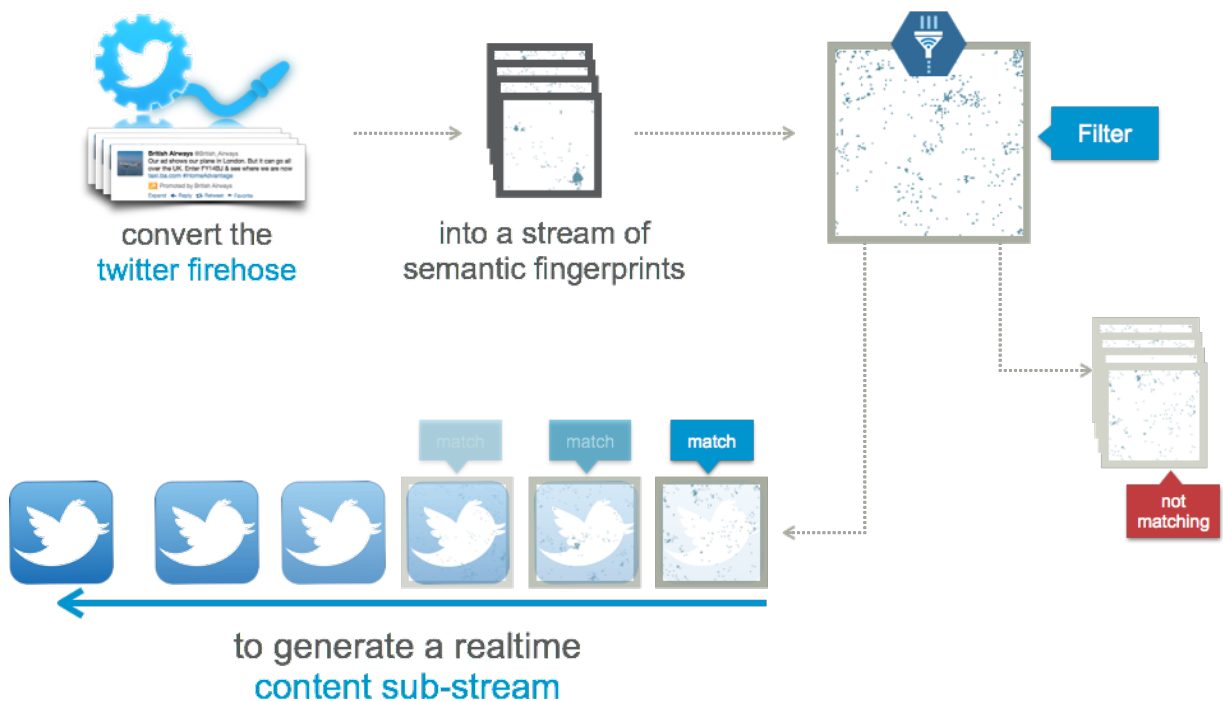

convert the
twitter firehose

into a stream of
semantic fingerprints

Filter

not
matching

match    match    match

to generate a realtime
content sub-stream

Fig. 18 Filtering the Twitter fire hose in real-time

Since the Semantic Fingerprint comparison process is extremely efficient, the content filtering can easily keep up with high frequency sources like the Twitter fire hose in real-time, even on very moderate hardware.

## Searching Documents

Using document similarity for enterprise search has been on the agenda of many products and solutions in the field. Widespread use of the approach has not been reached mainly because, lacking an adequate document (text) representation, no distance measures could be developed that could keep-up with the more common statistical search models.

With the Retina engine, searching is reduced to the task of comparing the fingerprints of all stored (indexed) documents with a query fingerprint that has either been generated by an example document ("Show me other documents like this one") or by typing in a description of what to look for ("Acts of vengeance of medieval kings").

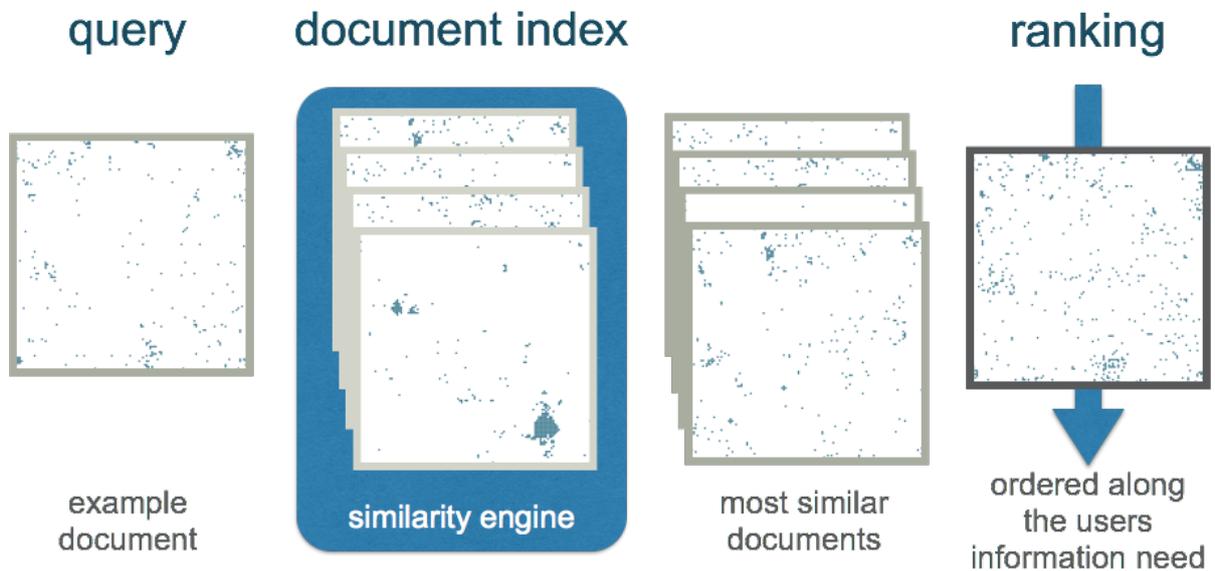

After the query fingerprint is generated, the documents are ordered by increasing distance. In contrast to traditional search engines, where a separate ranking procedure needs to be defined, the fingerprint-based search process generates an intrinsic order for the result set. Additionally, it is possible to provide personalized results by simply allowing the user to specify two or three documents that relate to his/her interests or working domain (without needing to be directly related to the actual search query). These user-selected domain documents are used to create a *user-profile-fingerprint*. Now the query is again executed and the (for example) 100 most similar documents are selected and are now sorted by increasing distance from the profile-fingerprint. Like this, two different users can cast the same search query on the same document collection and get different results depending on their topical preferences.

### Real-Time Processing Option

The Retina API has been implemented as an Apache Spark module to enable its use within the Cloudera infrastructure. This makes it possible to smoothly handle large text data loads of several terabytes and potentially even petabytes.

The ability to distribute fingerprint creation, comparison etc. across an arbitrarily large cluster of machines makes it possible to do real-time processing of data streams in order to immediately send a trigger, if some specific semantic constellations occur.

Document collections of any size can be classified, simplifying the application of Big Data approaches to unstructured text by orders of magnitude.

The efficiency of the Retina API combined with the workload aggregation capability of large clusters brings *index free searching* for the first time within reach of real-world datasets. By just implementing a *brute force* comparison of all document fingerprints with the query fingerprint, an index creation is not needed anymore. Most of the costs in maintenance and IT infrastructure related to large search systems originate from the creation, updating and management operations on the index.

In an index-free search system, any previously stored document can be found within microseconds.

## Using the Retina API with an HTM Backend

As stated in the beginning, HTM and Semantic Folding share the same theoretical foundations. All functionality described so far is solely based on taking advantage of the conversion of text into a SDR format.

In the following, the combination of the Retina API as text-data encoder with the HTM backend as *sequence learner* is used in a Text Anomaly Detection configuration.

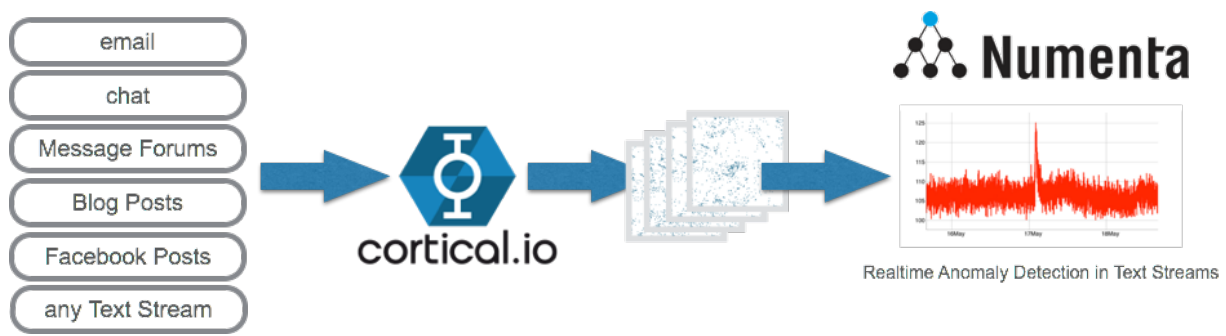

Fig.  20 Text Anomaly Detection using a HTM backend

Being exposed to a stream of text-SDRs, the HTM network learns word transitions and combinations that occur in real world sources. Based on the (text) data it was exposed to, the system constantly predicts what (word) it expects next. If the word it predicted is semantically sufficiently close to the word actually seen, the transition is strengthened in the HTM. If, on the other hand, an unexpected word (having a large semantic distance) occurs, an anomaly signal is generated.

## Advantages of the Retina API Approach

### Simplicity

1. No Natural Language Processing skills are needed.
2. Training of the system is fully unsupervised (no human work needed).
3. Tuning of the system is purely data driven and only requires domain experts and no specialized technical staff.
4. The API provided is very simple and intuitive to utilize.
5. The technology can be easily integrated into larger systems by incorporating the API over REST or by the inclusion of a plain Java library with no external dependencies for local (Cloudera/Apache Spark) deployments.

### Quality

1. Rich semantic feature set of 16K features allows a fine-grained representation of concepts.
2. All semantic features are self-learned, thus reducing semantic bias in the language model used.
3. The descriptive features are explicit and semantically grounded and can be inspected for the interpretation of any generated results.
4. By drastically reducing the vocabulary mismatch, far less false positive results are generated.

### Speed

1. Encoding the semantics in binary form (instead of the usual floating point matrices) provides orders of magnitudes of speed improvement over traditional methods.
2. All Semantic Fingerprints have the same size, which allows for an optimal processing pipeline implementation.

3. The semantic representations are pre-calculated and therefore don't affect the query response time.

4. The algorithms only apply independent calculations (no corpus relative computation) and are therefore easily scale to any performance needed.

5. The similarity algorithm can be easily implemented in hardware (FPGA & Gate Array technology) to achieve even further performance improvements. In a document search context, the specialized hardware could provide a stable query response time of < 5 microseconds, independently of the size of the searched collection.

### Cross-Language Ability

If aligned semantic spaces for different languages are used, the resulting fingerprints become language independent.

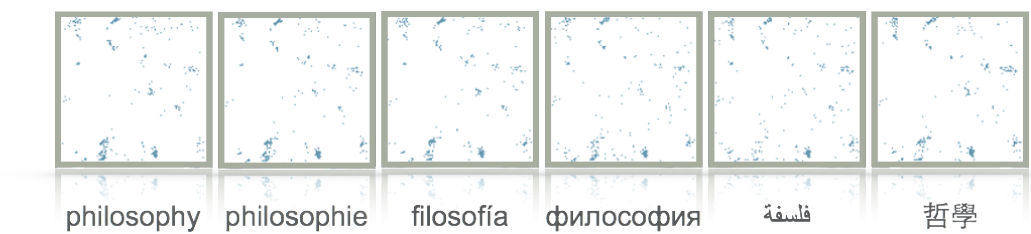

*Fig. 21 Language independence of Semantic Fingerprints*

This means that an English message-fingerprint can be directly matched with an Arabic message-fingerprint. When filtering text sources, the filter criterion can be designed in English while being directly applied to all other languages. An application can be developed using the English Retina while being deployed with a Chinese one.

## Outlook

A Retina System can be used wherever language models are used in traditional NLP systems. Upcoming experimental work will show if using a Retina system could improve Speech to Text, OCR or Statistical Machine Translation systems as they all generate candidate sentences from which they have to choose the final response by taking the semantic context into account.

Another active field of research is to find out if numeric measurements could also be interpreted as semantic entities like words. In this case the semantic grounding is not done by folding a collection of reference texts into the representation but by using log files of historic measurements. The correlation of the measurements will follow system specific dependencies as the correlation of words follow linguistic relationships and the system represented by the semantic space will not be "language" but an actual physical system from which the sensor data has been gathered.

A third field of research is to develop a hardware architecture that could speed-up the process of similarity computation. In very large semantic search systems, holding billions of documents, the similarity computation is the only remaining bottleneck. By using a content addressable memory (CAM) mechanism, the search-by-semantic-similarity-process could reach very high velocities.

# Part 3: Combining the Retina API with HTM

## Introduction

It is an old dream of computer scientists to make the meaning of human language accessible to computer programs. However, to date, all approaches based on linguistics, statistics or probability calculus have failed to come close to the sophistication of humans in mastering the irregularities, ambiguities and combinatorial explosions typically encountered in natural language.

Considering this fact, imagine the following experimental setup:

### Experimental Setup

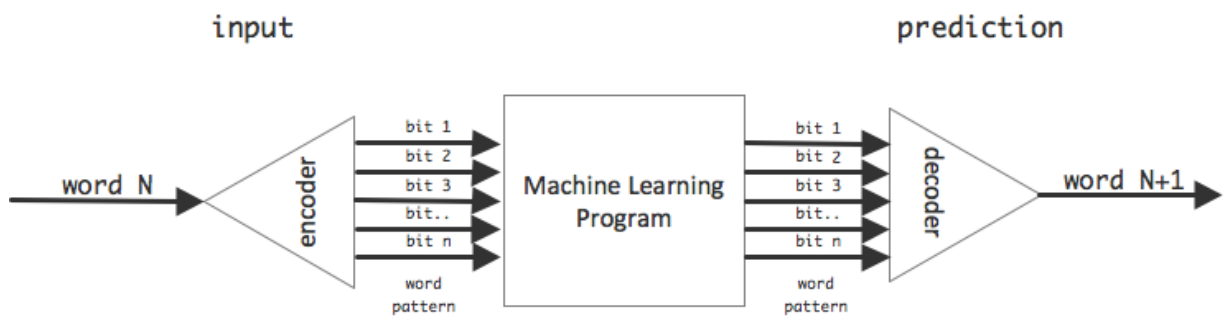

Fig. 22 Overview of the experimental setup

- A Machine Learning (ML) program that has a number of binary inputs. This ML program can be trained on sequences of binary patterns by exposing them in a time series. The ML program has predictive outputs that try to anticipate what pattern to expect next, in response to a specific anterior sequence.

- A codec program that encodes an English word into a binary pattern and decodes any binary pattern into the closest possible English word. The codec has the characteristic of converting semantically close words into similar binary patterns and vice versa. The degree of similarity between two binary patterns is measured using a distance metric such as Euclidian distance.

- The codec operates using a data width of 16Kbit (16384 bits) so that every English word is encoded into a 16Kbit pattern (binary word vector)

- The ML program is configured to allow patterns of 16Kbit as input as well as 16Kbit wide prediction output patterns.

- The codec is linked with the ML program to form a compound system that allows for words as input and words as output.

- The encoder part of the codec (word to pattern converter) is linked to the ML program inputs in order to be able to feed in sequences of words. After every word of a sequence, the ML program outputs a binary pattern corresponding to a prediction of what it expects next. The ML program grounds its predictions on the (learned) experience of sequences it had seen previously.

- The decoder part of the codec (pattern to word converter) is linked to the prediction outputs of the ML program. In this way, a series of words can be fed into the compound system that predicts the next expected word at its output based on previously seen sequences.

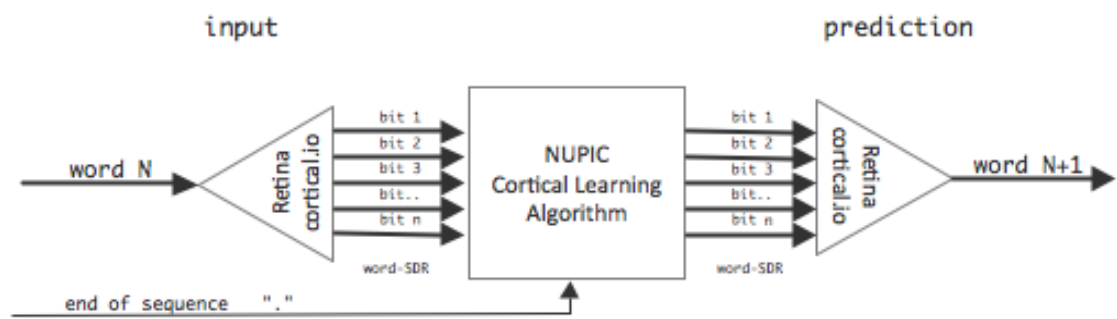

Fig. 23: Concrete experiment implementation

The ML program used in this experiment is the Hierarchical Temporal Memory (HTM) developed by Numenta. The code is publicly available under the name of NuPIC and actively supported by a growing community[iv]. NuPIC implements the cortical theory developed by Jeff Hawkins.

NuPIC is a Pattern Sequence Learner. This means that the initially agnostic program[6] can be trained on sequences of data patterns and is able to predict a next pattern based on previously exposed sequences of patterns.

In the following experiments the data consists of English natural language. We use the Cortical.io API to encode words into binary patterns, which can be directly fed into the HTM Learning Algorithm (HTM LA). Being an online learning algorithm, the HTM LA learns every time it is exposed to input data. It is configured to store frequently occurring patterns and the sequences they appear in.

---

[6] In its initial state the algorithm does not know of any SDR or sequence thereof.

After the HTM LA has read a certain number of words, it should start to predict the next word depending on the words read previously. The learning algorithm outputs a binary prediction pattern of the same size as the input pattern, which is then decoded by the Cortical.io API back into a word[7].

The *full stop* at the end of a sentence is interpreted by the HTM LA as an *end-of-sequence* signal, which ensures that a new sequence is started for each new sentence.

### Experiment 1: "What does the fox eat?"

In this first experiment, the setup is used in the simplest form. A dataset of 36 sentences, each consisting of a simple statement about animals and what they eat or like, is fed in sequence into the HTM LA. A new pattern sequence is started after each full stop by signaling it to the HTM LA. Each sentence is submitted only once. The HTM LA sees a binary pattern of 16K bits for each word and does not know anything about the input, not even which language is being used.

---

[7] The Cortical.io API returns the word for the closest matching fingerprint.

## Dataset

The following 36 sentences are presented to the system:

```
1.  frog eat flies.            19. cow eat grain.
2.  cow eat grain.             20. elephant eat leaves.
3.  elephant eat leaves.       21. goat eat grass.
4.  goat eat grass.            22. wolf eat rabbit.
5.  wolf eat rabbit.           23. sheep eat grass.
6.  cat likes ball.            24. cat eat salmon.
7.  elephant likes water.      25. wolf eat mice.
8.  sheep eat grass.           26. lion eat cow.
9.  cat eat salmon.            27. coyote eat mice.
10. wolf eat mice.             28. elephant likes water.
11. lion eat cow.              29. cat likes ball.
12. dog likes sleep.           30. coyote eat rodent.
13. coyote eat mice.           31. coyote eat rabbit.
14. coyote eat rodent.         32. wolf eat squirrel.
15. coyote eat rabbit.         33. dog likes sleep.
16. wolf eat squirrel.         34. cat eat salmon.
17. cow eat grass.             35. cat likes ball.
18. frog eat flies.            36. cow eat grass.
```

*Please note that, for reasons of simplicity, the sentences are not necessarily grammatically correct.*

Fig. 24: The "What does the fox eat" experiment

## Results

The HTM LA is a so-called Online Learning System that learns whenever it gets data as input and has no specific training mode. After each presented word (pattern), the HTM LA outputs its best guess of what it expects the next word to be. The quality of predictions rises while the 36 sentences are learned. We discard these preliminary outputs and only query the system by presenting the beginning of a 37th sentence "fox eat". The final word is left out and the prediction after the word *eat* is considered to be the answer to the implicit question:

**"what does the fox eat?".** The system outputs the word ***rodent.***

## Discussion

The result obtained is remarkable, as it is *correct* in the sense that the response is actually something that could be food for some animal, and also *correct* in the sense that rodents are actually typical prey for foxes.

Without knowing more details, it looks as if the HTM was able to understand the meaning of the training sentences and was able to infer a plausible answer for a question about an animal that was not part of its training set. Furthermore, the HTM did not pick the correct answer from a list of possible answers but actually *synthesized* a binary pattern, for which the closest matching word in the Cortical.io Retina[8] happens to be *rodent*.

## Experiment 2: "The Physicists"

The second experiment uses the same setup as in experiment 1. This time, a different set of training sentences is used. In the first case it was the goal to generate a simple inference based on a single list of examples. Now, the inference is structured in a slightly more complex fashion. The system is trained on examples of two different professions: physicists and singers and what they like (mathematics and fans) and what the profession actors like (fans).

---

[8] This Retina has been trained on 400K Wikipedia pages. This is also the reason why it could understand (by analogy) what a fox is without ever having seen it in the training material. What it has been seeing are the words *wolf* and *coyote*, which share many bits with the word *fox*.

## Dataset

Physicists:

1. marie curie be physicist.
2. hans bethe be physicist.
3. peter debye be physicist.
4. otto stern be physicist.
5. pascual jordan be physicist.
6. felix bloch be physicist.
7. max planck be physicist.
8. richard feynman be physicist.
9. arnold sommerfeld be physicist.
10. enrico fermi be physicist.
11. lev landau be physicist.
12. steven weinberg be physicist.
13. james franck be physicist.
14. karl weierstrass be physicist.
15. hermann von helmholtz be physicist.
16. paul dirac be physicist.

What Physicists like:

1. eugene wigner like mathematics.
2. wolfgang pauli like mathematics.

What Actors like:

1. pamela anderson like fans.
2. tom hanks like fans.
3. charlize theron like fans.

Singers:

1. madonna be singer.
2. rihanna be singer.
3. cher be singer.
4. madonna be singer.
5. elton john be singer.
6. kurt cobain be singer.
7. stevie wonder be singer.
8. rod stewart be singer.
9. diana ross be singer.
10. marvin gaye be singer.
11. aretha franklin be singer.
12. bonnie tyler be singer.
13. elvis presley be singer.
14. jackson browne be singer.
15. johnny cash be singer.
16. linda ronstadt be singer.
17. tina turner be singer.
18. joe cocker be singer.
19. chaka khan be singer.
20. eric clapton be singer.
21. elton john be singer.
22. willie nelson be singer.
23. hank williams be singer.
24. mariah carey be singer.
25. ray charles be singer.
26. chuck berry be singer.
27. cher be singer.
28. alicia keys be singer.
29. bryan ferry be singer.
30. dusty springfield be singer.
31. donna summer be singer.
32. james taylor be singer.
33. james brown be singer.
34. carole king be singer.
35. buddy holly be singer.
36. bruce springsteen be singer.
37. dolly parton be singer.
38. otis redding be singer.
39. meat loaf be singer.
40. phil collins be singer.
41. pete townshend be singer.
42. roy orbison be singer.
43. jerry lee lewis be singer.
44. celine dion be singer.
45. alison krauss be singer.

What Singers like:

1. katy perry like fans.
2. nancy sinatra like fans.

*Please note that, for reasons of simplicity, the sentences are not necessarily grammatically correct.*

**Fig. 25: The "The Physicists" experiment**

## Results

The program is started using the datasets above. This is the terminal log of the running experiment:

```
Starting training of CLA ...

.  .  .  .  .  .  .  .  .  .  .  .

.  .  .  .  .  .  .  .  .  .  .  .  .  .  .
.  .  .  .  .  .  .  .  .  .  .  .  .  .  .
.  .  .  .  .

.  .

.  .

.  .  .
Finished training the CLA.
Querying the CLA:
eminem be => singer
eminem like => fans
niels bohr be => physicist
niels bohr like => mathematics
albert einstein be => physicist
albert einstein like => mathematics
tom cruise like => fans
angelina jolie like => fans
brad pitt like => fans
physicists like => mathematics
mathematicians like => mathematics
actors like => fans
physicists be => physicist
```

Fig. 26: Terminal log showing the results of "The Physicists" experiment

## Discussion

After training the HTM with this small set of examples (note that the classes "What Physicists like" and "What Singers like" are only characterized by two example sentences), a set of queries based on unseen examples of singers, actors and physicists is submitted. In all cases, the system was able to make the correct inferences, regardless of the verb used (be, like). The last four queries suggest that the system was also able to generalize from the concrete examples of the training sentences towards the corresponding class labels like *physicists*, *actors* and *singers* and associate to them the correct *like-preferences*.

For these inferences to be possible, the system has to have access to some real world information. As the HTM itself had no preemptive knowledge, the only possible source for bringing in this information would have been through the language elements used as training material. But the very small amount of training material clearly does not contain all that background in a descriptive or declarative form. So the only point where the relevant context could have been introduced is through the encoding step, converting the symbolic string into a binary word pattern.

# Part 4: References

## Web

## Videos

**Analogy as the Core of Cognition,** Douglas Hofstaedter, Stanford University, 2009, retrieved from https://youtu.be/n8m7lFQ3njk

**Linguistics as a Window to Understanding the Brain,** Steven Pinker, Harvard University, 2012, retrieved from https://youtu.be/Q-B_ONJIEcE

**Semantics, Models, and Model Free Methods**, Monica Anderson, Syntience Inc., 2012, retrieved from https://vimeo.com/43890931

**Modeling Data Streams Using Sparse Distributed Representations,** Jeff Hawkins, Numenta, 2012, retrieved from https://youtu.be/iNMbsvK8Q8Y

**Seminar: Brains and Their Application,** Richard Granger, Thayer School of Engineering at Dartmouth, 2009**,** retrieved from https://youtu.be/AsYOl4REct0

## Cortical.io

**Website** http://www.cortical.io

**Retina API** http://api.cortical.io

**Demos**
Keyword Extraction http://www.cortical.io/keyword-extraction.html

Language Detection http://www.cortical.io/language-detection.html

Expression Builder http://www.cortical.io/expression-builder.html

Similarity Explorer http://www.cortical.io/demos/similarity-explorer/

Topic Explorer Demo http://www.cortical.io/topic-explorer.html

Cross-Lingual Topic Analyzer http://www.cortical.io/cross-language.html

Topic Modeller http://www.cortical.io/static/demos/html/topic-modeller.html